# An Ethical Trajectory Planning Algorithm for Autonomous Vehicles

Maximilian Geisslinger[1✉], Franziska Poszler[2] and Markus Lienkamp[1]

**With the rise of AI and automation, moral decisions are being put into the hands of algorithms that were formerly the preserve of humans. In autonomous driving, a variety of such decisions with ethical implications are made by algorithms for behavior and trajectory planning. Therefore, we present an ethical trajectory planning algorithm with a framework that aims at a fair distribution of risk among road users. Our implementation incorporates a combination of five essential ethical principles: minimization of the overall risk, priority for the worst-off, equal treatment of people, responsibility, and maximum acceptable risk. To the best of the authors' knowledge, this is the first ethical algorithm for trajectory planning of autonomous vehicles in line with the 20 recommendations from the EU Commission expert group and with general applicability to various traffic situations. We showcase the ethical behavior of our algorithm in selected scenarios and provide an empirical analysis of the ethical principles in 2000 scenarios. The code used in this research is available as open-source software.**

Researchers, as well as industry, around the world, are working on the development of autonomous vehicles (AVs). Therefore, the task of a human driver is being replaced by sophisticated sensors and actuators with software at the center of development. However, scholars and experts admonish that replacing human beings also has to consider ethical aspects[1,2]. In particular, accidents involving autonomous vehicles[3] raise not only technical questions but also have ethical implications. They are also important for potential adopters of AVs.[4] To move ahead in the development of AVs, that satisfy ethical requirements, we have to consider ethical aspects of the AV's software. As the software part responsible for decision-making, behavioral and trajectory planning plays a decisive role in this process.

In the past, to integrate ethics into software algorithms, scholars have investigated traditional ethical frameworks such as deontology[5] or consequentialism[6]. Recent works advocate considering aspects of risk ethics in autonomous vehicle decision-making [7] as well as integrated approaches[8]. The associated advantages relate to the consideration of uncertainties through probabilities[9, 10], the given explainability [11], and scalability to a wide variety of driving situations and not only dilemma situations[11]. The consideration of risk shifts questionable life-and-death decisions to the question of who is put at what risk.[12] This argumentation is supported by the Horizon EU Commission Expert Group[13], which considers dilemmas as a limit case of risk management. Their recommendations suggest the use of "shared ethical principles in risk management" [13]. The German Ethics Code[14] for Autonomous Driving also supports this approach by arguing for "balancing risks".

Motion planning for AVs has a strong impact on risk distribution among road users, at least implicitly or even explicitly[15]. According to a study[16], passengers of AVs have a personal incentive to ride in AVs that will protect them at all costs. This incentivizes developers and OEMs to program AVs with an appropriate selfish risk distribution towards the AV passengers. This type of risk distribution could ultimately lead to weaker road users being disadvantaged by higher risks[7], as we will show in our results. For this reason, derived from ethical theroies, we argue for a fair distribution of risk for all road users in traffic, as implemented with our proposed algorithm.

At present, to the best of the authors' knowledge, there is no algorithm that allows for a fair distribution of risks and builds on the high-level ethical considerations and recommendations from the EU. In this paper, we provide a link between ethical foundations and a real trajectory planning algorithm. Therefore, we provide a top-down ethical framework for trajectory planning in line with the EU that leaves space for bottom-up modifications. In this way, we meet the basic ethical requirements, but leave room for modifications that may be necessary due to different cultures[17] or areas of application, or even for each individual[18,19]. However, to what extent such a cultural determination is admissible would be subject to further research.

[1] Institute of Automotive Technology, Technical University of Munich, Germany  
[2] Institute for Ethics in Artificial Intelligence, Technical University of Munich, Germany  
✉ maximilian.geisslinger@tum.de

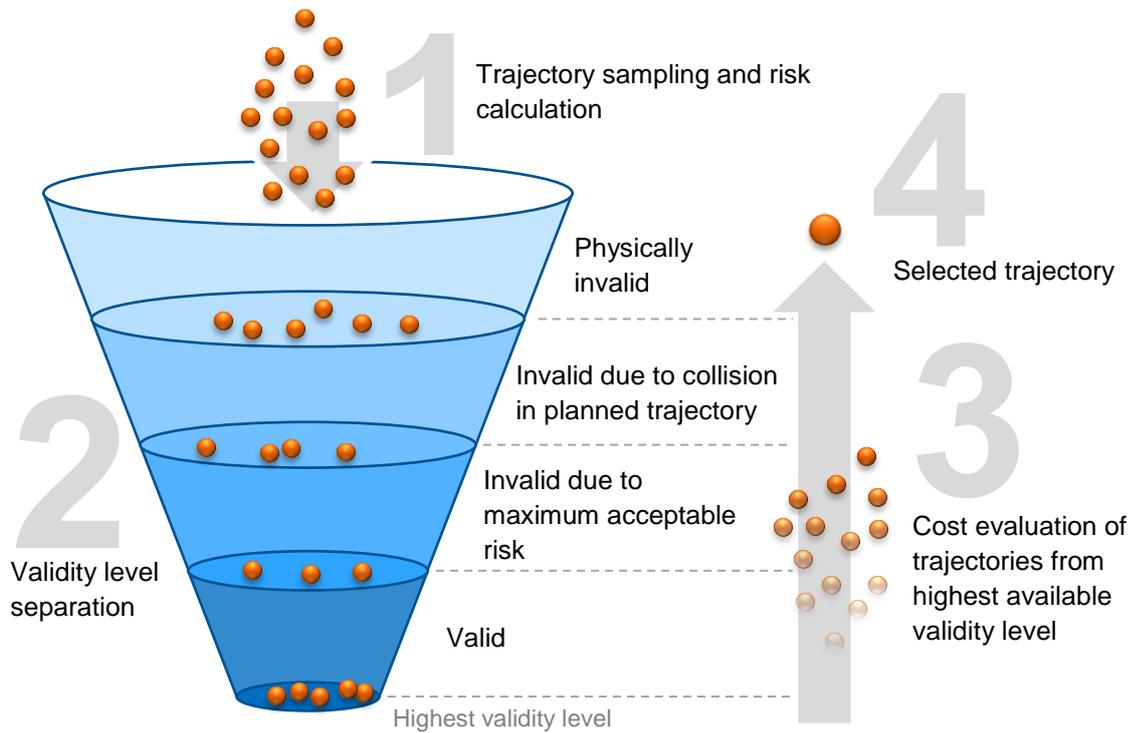

**Fig. 1 | Our ethical trajectory planning algorithm in four steps.** The small orange balls symbolize various sampled trajectories, of which one is chosen to be executed.

## The Algorithm

Our software for AV behavior and trajectory planning consists of four steps that we will highlight in this paper. The overall structure of the algorithm, as shown in Fig. 1, allows for ethical decision-making. Therefore, we will quickly summarize this algorithm structure to explain how we integrate ethics later in detail.

1. According to a pre-specified discretization scheme, multiple trajectories are sampled given the current state of the AV. Various methods for generating the trajectories are possible here. In this work, we generate the trajectories in a Frenet coordinate system [20]. In addition, we assign a risk value for every road user for every trajectory, which allows consideration of risk distributions in trajectory planning as a key aspect of ethical decision-making.
   ( ▶ Section Risk Calculation)

2. The sampled trajectories are categorized into four validity levels. While we always aim to have trajectories at the highest validity level, this ensures that a solution is found even if not all quality requirements are met. From an ethical point of view, we actively consider a risk threshold here, which we denote as the maximum acceptable risk.
   ( ▶ Section Maximum Acceptable Risk)

3. For those trajectories within the highest available validity level (e.g. "valid"), an ethical cost function is calculated. We use different cost functions depending on the validity level. Among other critical dimensions, the cost function ensures a fair distribution of risk. We will illuminate various ethical aspects that are considered.
   ( ▶ Section Risk Distribution)

4. The trajectory from the highest available validity level and the lowest calculated cost is selected and executed. After a single time step is completed, a new trajectory is planned for the next time step starting from step 1.

## Risk Calculation

In our work, we adopt the definition of risk as an expected value. Accordingly, risk is defined as the product of a probability that a certain event will occur and a measure of the consequences of that event[21]. In the case of autonomous driving, we can therefore define the risk as the product of a collision probability $p$ and the estimated harm out of that collision $H$[7]. Driving autonomous vehicles on the road inevitably requires accepting a certain residual risk for collisions[22, 23]. Otherwise, they are subject to the so-called freezing robot problem[24].

Probabilities for potential collisions result from different kinds of uncertainties in autonomous driving. These uncertainties range from environmental perception, such as detection and localization of other road users, to uncertainties in vehicle control. In this work, we focus on one of the most significant causes of uncertainty in trajectory planning: the prediction of other road users' trajectories. We use a neural network model[25], which outputs probability-based trajectory predictions as a bivariate normal distribution around a most-likely trajectory prediction. Thus, we calculate a collision probability for every planned ego-state given the object's shapes.

While calculating collision probabilities as a result of prediction uncertainties is mainly a technical task, the estimation of harm as the second part of risk management raises ethical questions. Harm modeling aims to map the estimated personal damage to a scale of values from 0 to 1, where 0 denotes a collision without any harm (to humans) and 1 accounts for the greatest possible harm. Modeling harm in advance is difficult and is a field of research on its own. For ethical trajectory planning, we need an estimate with the capability of fast calculation at runtime. Also, from an ethical point of view, we want to expressly exclude certain factors from our analysis, which simplifies the model. For example, we deliberately do not intend to make any judgment as to which physical harm is more severe: whether death is the highest possible damage or whether a life-long limiting injury prevails is explicitly not to be decided in general. Therefore, we choose a degree of abstraction in the model by establishing universally valid relations between the available input variables and the severity of the damage. For example, a proportional relationship between the velocities of the traffic participants and the severity of harm can be established independently of any ethical evaluation. Other impact factors that we consider are the mass of both colliding parties, the angle of impact, and the impact area.

Moreover, the damage calculated on this basis does not account for any property damage but refers to the expected personal injury, which must be clearly prioritized in any case. We also do not consider any personal characteristics in our model, such as age, gender, or quality-of-life (QOL)[26] measures. In this way, we also comply with established requirements of the ethics committees and the basic legal situation in many countries.

In line with the recommendations of ethics committees[27], our model distinguishes between protected (vehicles, trucks, etc.) and unprotected (pedestrians, cyclists, etc.) road users. For both cases, we run a logistic regression using the NHTSA's Crash Report Sampling System (CRSS)[28]. This results in Equations (1) and (2), where $m$ and $v$ are the mass and velocity of the two road users A and B, $\alpha$ is the collision angle and $c_0$, $c_1$ and $c_{\text{area}}$ are empirically determined coefficients. The probability of an accident with severity MAIS3+ serves as a comparison value to map the damage to a value between 0 and 1. We mitigate bias against co-drivers through that database by aligning the impact areas symmetrically.

$$\Delta v_A = \frac{m_B}{m_A + m_B} \sqrt{v_A^2 + v_B^2 - 2\, v_A v_B \cos \alpha} \qquad (1)$$

$$H = \frac{1}{1 + e^{c_0 - c_1\, \Delta v - c_{\text{area}}}} \qquad (2)$$

Using the harm model, we assign estimated harm to every collision probability. Thus, we know the time-variant risk along the planning horizon for every sampled trajectory. However, we want to describe every possible trajectory by a single risk value. This becomes difficult as the collision probabilities along the planning horizon and thus the risks for a collision of the same road users are not independent. Therefore, we describe the risk $R_{\text{Traj, dep.}}$ of a trajectory $u$ with the maximum risk over time:

$$R_{\text{Traj, dep.}}(u) = \max\bigl(p(u)\, H(u)\bigr) \qquad (3)$$

This allows the assignment of a risk value for every road user depending on the planned trajectory of the AV, as shown in Fig. 2.

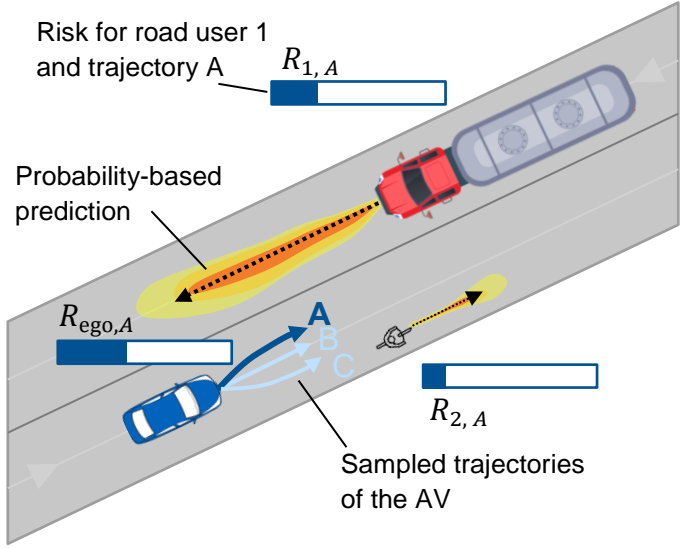

**Fig. 2 | Trajectory planning based on risk distribution.** On the basis of a probability-based prediction of all road users and an estimated harm value, every trajectory of the AV can be assigned risk values for every road user.

Risks that originate from different collisions are assumed as being independent and thus calculated with Equation (4), where $S_R$ denotes a set of risks for all road users:

$$R_{\text{Traj, indep.}} = 1 - \prod_{|S_R|} \left(1 - R_{\text{Traj, dep.}}\right) \quad (4)$$

We calculate a combined risk from possible collisions with multiple road users only for the ego-vehicle in the case of the maximum acceptable risk principle for reasons of information asymmetry.

## Maximum Acceptable Risk

The preceding quantification of risks prompts the question of what amount risk may be permitted as a maximum. Today, in road traffic, risk thresholds are implicitly set by authorities and state regulations: a speed limit within urban areas, or regulations on safety distances, implicitly represent a weighting of acceptable risk versus values such as traffic flow for all road users. Along with this, registration of AVs faces explicit questions about how safe is safe enough. The United Nations Economic Commission for Europa (UNECE) states that an AV "shall not cause any non-tolerable risk".[29] In our framework, we introduce a value called *maximum acceptable risk* $R_{\max}$. This risk threshold denotes which trajectory is allowed and which one is only allowed if there is no other option. We provide validity levels in our algorithm, where the selected trajectory must always be selected from the highest available validity level. Only if no other options are possible, can an alternative trajectory be planned, which for example, exceeds $R_{\max}$. This leads to a decision framework on a maneuver basis. As shown in Fig. 3, this method can be used to decide whether a potentially dangerous overtaking maneuver should be performed or not. Thus, the determination of the value is not only in the interest of AV users but in the interest of all road users, as shown by the scenario.

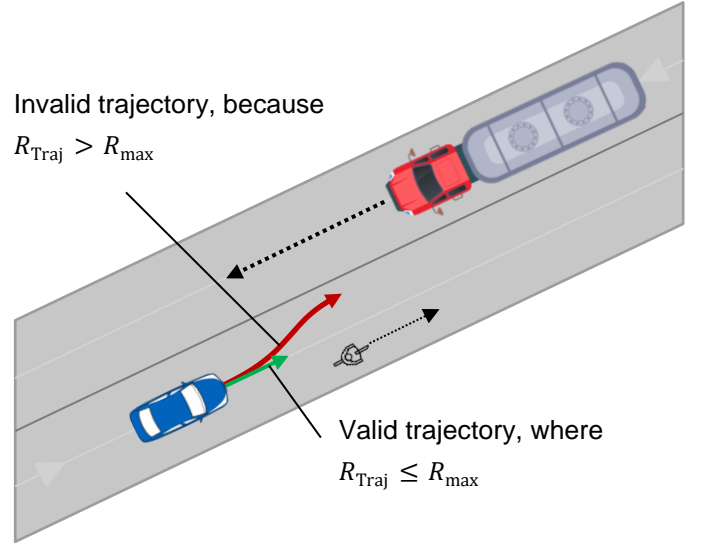

**Fig. 3 | Visualization of the maximum acceptable risk principle.** The trajectory to overtake the cyclists yields too high a risk value and is not allowed, as there is a valid alternative (staying behind the cyclist).

Working with this kind of risk threshold creates, on the one hand, better decisions under the assumption of a reliable risk assessment of AVs and, on the other hand, transparency towards all traffic participants concering the AV's functioning. It enables social debate on an important question within our society:

> *"How much risk are we as a society willing to accept in exchange for mobility?"*

This question goes far beyond the issue of autonomous vehicles. It is likely to arise in more and more areas where understanding of risk through modern algorithms and big data is growing.

An initial indication of such a value in road traffic is provided by the current safety level with human drivers. However, it must be considered that a significantly higher level of safety must be ensured for widespread acceptance due to various effects, such as the psychological mechanisms of AV adopters.[30]

## Risk Distribution

A cost function generally regulates different target variables for trajectory planning. Essentially, three goals can be distinguished, which must be weighed by parameterization: *mobility, safety (or risk), and comfort*. Mobility pursues the goal of reaching a destination as quickly as possible and is often in direct contrast to safety, which favors lower velocities. Comfort is taken into account in the cost function by quantities such as acceleration and jerk and thus plays a subordinate but not negligible role.

Our algorithm uses different cost functions depending on the highest available validity level $l_v$. If the trajectories to choose from are *valid*, which should usually be the case, then the cost function considers all three target variables. Hence, the function for calculating the total trajectory costs $J_\text{total}$ is a parameterized combination of costs for risk, mobility, and comfort:

$$J_\text{total}(u \mid l_v = \text{valid}) \\ = J_\text{Risk}(u) + J_\text{Comfort}(u) \\ + J_\text{Mobility}(u) \quad (5)$$

Suppose a valid trajectory can no longer be generated, for example, because the maximum acceptable risk does not hold or no collision-free trajectory can be found. In that case, another cost function must be used. In line with ethics committees' requirements for safety as the highest objective[27], the risk part must be prioritized higher than mobility and comfort because of a possible accident threat. Consequently, we only consider the risk part until valid trajectories are available again.

$$J_\text{total}(u \mid l_v \neq \text{valid}) = J_\text{Risk}(u) \quad (6)$$

The risk cost function $J_\text{Risk}$ consists of four ethical principles guiding risk distribution. As stated by the EU Commission[13] and previous work[7], a single principle for risk distribution does not satisfy the requirements for ethical choices. We will discuss four principles and argue why they are needed for fair risk distribution in road traffic.

Translating ethical distribution principles into a mathematical representation in the form of trajectory costs provides the basis for our ethical algorithm. In a weighted sum, our risk cost function combines costs for the Bayes principle $J_B$, the equality principle $J_E$, maximin principle $J_M$, and the responsibility principle $J_R$. The associated weighting factors $w$ determine how strongly the respective principle is to be taken into account. Note that $J_R$ has a negative sign and the same weight $w_B$ as the Bayes principle. This results in less risk consideration of road users responsible for the risk in the cost function. At the same time, using $w_B$ here ensures that the algorithm never strives to increase the risks of responsible road users without any benefit.

$$J_\text{Risk}(u) = w_B J_B(u) + w_E J_E(u) + w_M J_M \\ - w_B J_R(u) \quad (7)$$

## Bayes, Equality & Maximin Principle

The Bayes principle in risk ethics follows a utilitarian approach and corresponds to a minimization of the overall risk. For this purpose, the risks of all detected road users in a scenario are cumulated and normalized by the number of road users according to Equation (8). As with the other principles, we design the range of values to be independent of the number of road users. Otherwise, the risks would shift accordingly with more road users and a universal distribution would no longer be valid. According to Bayes' principle, the trajectory with the lowest overall risk must be selected in the cost function. This principle ensures that the AV makes an optimal decision for all road users as a whole. Therefore, this principle is also preferred by a majority of road users[31].

$$J_B(u) = \frac{\sum_{i=1}^{|S_R|} R_i(u)}{|S_R|} \quad (8)$$

However, the Bayes principle only minimizes the overall risk and does not account for fairness in risk distribution. As one aspect of fairness, we propose distributing risks equally among road users. We introduce the equality principle to ensure that using Bayes' principle does not create a bias with exceptionally high risks for individual road users in favor of a low overall risk. By summing up differences in risk, the principle described by Equation (9), pursues equal treatment of people. Normalization with the number of summation terms also allows a consideration of the costs independent of the number of road users.

$$J_E(u) = \frac{\sum_{i=1}^{|S_R|} \sum_{j=i}^{|S_R|} |R_i(u) - R_j(u)|}{\sum_{k}^{|S_R|-1} k} \quad (9)$$

The principles introduced so far consider risk as a whole and do not allow separate consideration of collision probability and harm. In this context, a high collision probability with low personal injury does not per se equate to a lower probability of a fatal accident. To enable decoupled assessment, we use the maximin principle, which goes back to Rawls' theory of justice[32]. Applied to the context of AVs, it demands minimizing the greatest possible harm regardless of probability. To achieve this, the principle demands the trajectory where the greatest possible harm, described with the costs in Equation (10), is minimal. This amounts to the idea of prioritizing the worst-off. Because this term neglects the usually low probabilities, we introduce a discount factor $\gamma \geq 1$.

$$J_M(u) = [max_{H_i(u)}(S_H)]^\gamma \quad (10)$$

## Responsibility Principle

Risks in road traffic arise from the characteristics or actions of road users. Therefore, these characteristics and actions can be attributed to a moral responsibility for the risks that arise from them. Consequently, the degree of moral responsibility that people bear for creating risky situations must be considered.[33] Recent approaches guarantee safety for AVs in terms of not being responsible for any accident[34]. This shows that the consideration of responsibility in the decision-making process plays an important role.

Considering moral responsibility not only accounts for fairness but also ensures that the human road users involved behave responsibly. Abusing the crash avoidance system of AVs by abruptly running in front of the autonomous vehicle as a pedestrian - knowing that the vehicle will stop – may be a future problem for AVs, which has already been discussed[1,23]. The consideration of responsibility in risk distribution can help to prevent such irresponsible behavior. This is also in line with the EU Commission who favors "creating a culture of responsibility"[13].

In the risk cost function, responsibility is taken into account by reducing the considered risk cost of responsible road users from the Bayes term. $R_i$ accounts for the risk of the corresponding traffic participant. $r_i$ describes the amount of responsibility, where a value of "0" stands for no cost reduction due to responsibility and "1" would reduce all the risk costs from the Bayes term for that road user. Note that, in practice, we do not allow $r_i$ to be 1, as in this case, our algorithm would not care about the risk of the responsible road user anymore at all.

$$J_R(u) = \sum r_i(u) R_i(u), r_i \in [0, 1) \quad (11)$$

There are three different types of responsibility, which we consider accordingly in our planning algorithm as indications of how much risk individuals bring to traffic: *the adherence to traffic rules*, *the characteristics of road users*, and *the space of actions of road users*.

Traffic rules ensure that road users can interact safely with each other so that accidents and the resulting damage can be avoided as far as possible. They form the basis for the fundamental idea that everyone can rely on the correct behavior of other road users if one behaves appropriately himself. Thus, it seems reasonable to account for any violations of traffic rules in our risk distribution. To calculate the responsibility of road users in terms of traffic rule adherence, we calculate the reachable set of every road user[35]. A reachable set calculates a time-variant area that can be physically and legally (according to traffic rules) reached by a road user[36]. If a possible collision is calculated inside a road user's reachable set, the road user would be considered as not responsible in terms of traffic rules for this collision. On the other hand, if a potential collision is outside of a reachable set, a collision is either not possible (because of the physical constraints) or the road user can be considered responsible, as shown in Fig. 4. Thus, they will be assigned responsibility for costs. The amount would depend on the kind and severity of the rule violation, which needs to be further investigated and deliberated in future research.

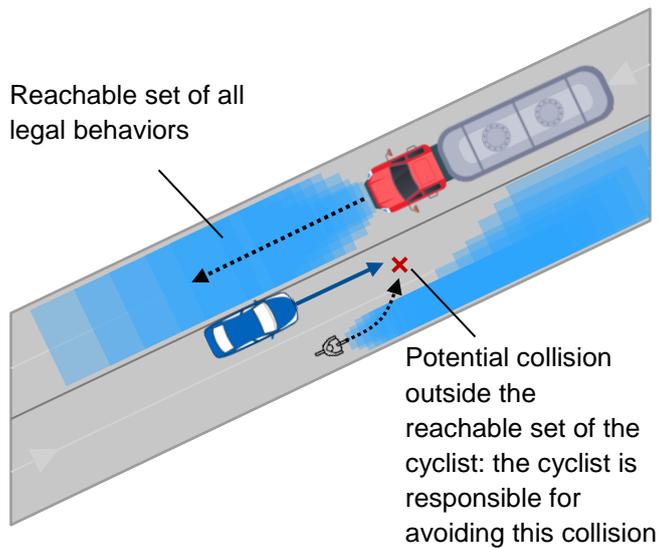

**Fig. 4 | Schematic example for the responsibility principle in the case of a rule violation.** A reachable set of all legal behaviors for all road users builds the basis for assigning responsibility.

In line with strict liability, the responsibility principle must also address the characteristics of different road users and their associated moral claims[37]. Vulnerable road users (VRUs), such as cyclists or pedestrians, introduce much less risk into road traffic than cars or trucks because of their mass and velocities. Without motorized road users, there would hardly be any traffic fatalities. For this reason, motorized participants also have a special responsibility to protect weaker road users. Algorithmically, this is implemented by assigning a (small) responsibility $r_i$ to motorized road users regardless of their behavior, but depending on their mass and the corresponding risk they bring into traffic.

The third kind of responsibility we consider in our ethical algorithm is due to the space of actions. Each vehicle has different possible options in trajectory planning. This may result in different responsibilities of each vehicle in possible collisions. Imagine a vehicle crashing into another vehicle from behind in a rear-end collision. The vehicle behind is responsible for this collision, as it is also handled, for example, in the settlement of claims. We incorporate these responsibilities into our trajectory planning using reachable sets.

## Results

At the beginning of this paper, we motivated the need for a fair (ethical) risk distribution in contrast to a selfish risk distribution in the AV trajectory planning. We subsequently presented an ethical trajectory planner with a fair distribution of risk, which we will empirically evaluate in this section. Therefore, we address whether the proposed ethical algorithm results in a fair distribution of risk. For this purpose, we compare our ethical algorithm with a selfish algorithm and a standard algorithm. The selfish algorithm strives to minimize the AV's risk and does not account for the risks of other road users, while the standard algorithm does not consider risk at all. For our empiric evaluation, we run these three types of algorithms in 2000 simulation scenarios using CommonRoad[38]. The scenarios are partly real-world recorded or handcrafted to create explicitly safety-critical situations. They cover a wide variety of environments, such as urban, rural, and highway, in different regions, such as the USA, Europe, and China. As the simulated traffic participants behave deterministically and do not interact with the AV, we do not consider the responsibility principle in our evaluation here.

Within the scope of this evaluation, we focus on two dimensions: on the one hand, we analyze the risks arising from the different planning approaches, and on the other hand, we analyze the simulated accidents that occur. For this purpose, we compare three groups of road users: (1) the AV itself (Ego-AV), (2) all other road users except the AV (3rd Party), and (3) explicitly, as a subgroup the vulnerable road users (VRUs). Fig. 5 shows the distribution of the 100 highest risk values appearing for the corresponding group of road users. We focus on the 100 highest values because most of the risk values in the scenarios are similarly close to zero with all three algorithms and the higher values are more relevant. The analysis indicates that explicitly considering risks in the trajectory planning reduces the risk for all road users. Comparing the selfish, algorithm with our proposed ethical approach, we observe similar risks for the ego vehicle as with the selfish algorithm. However, at the same time, risks for 3rd party road users, and especially VRUs, are significantly lower. So, using the ethical algorithm instead of a selfish one means shifting risk from VRUs to the ego vehicle, whereas the benefit for VRUs is higher than the downside for the ego vehicle. Still, the VRUs have to carry higher risk due to the higher harm they experience in the event of an accident.

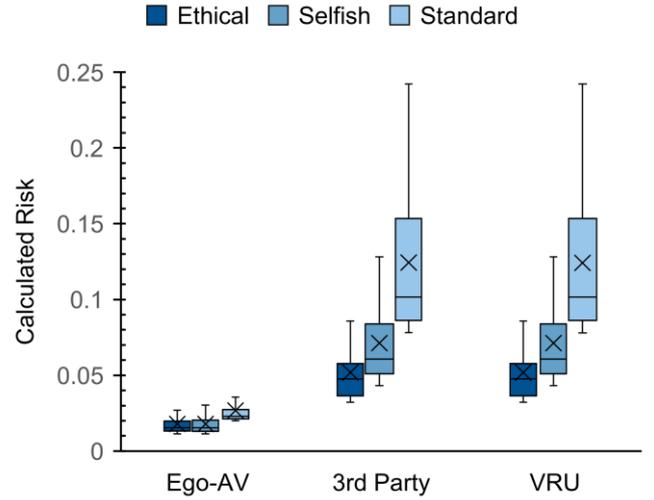

**Fig. 5 | Risk distribution of the highest 100 occurring risks.** Three different algorithms (ethical, selfish, and standard) are compared on three different groups of road users (Ego-AV, 3rd party, and VRU).

The examination of the accidents in the simulation and the associated harm confirms these findings. Fig. 6 shows the cumulated harms for the corresponding groups of road users as a result of the 2000 scenarios. Only in the case of 3rd party road users, are the VRUs no longer as significant as in Fig. 5. Thus, the calculated harm for the ethical and selfish algorithms are similar here. The discrepancies between the distribution of actual accident harm and the risks can be attributed, on the one hand, to unavoidable inaccuracies in the risk modeling and, on the other hand, to the small amount of data on accidents.

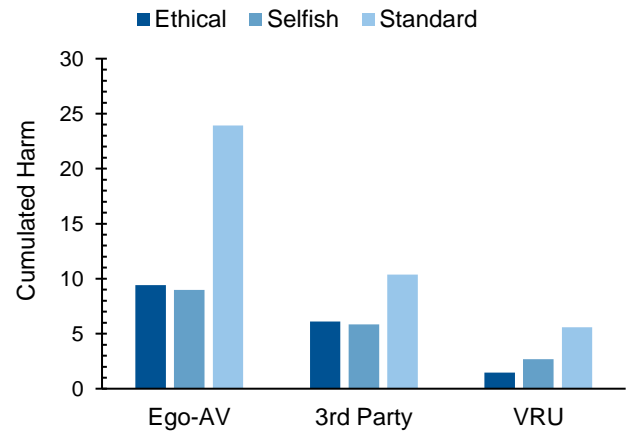

**Fig. 6 | Cumulated personal harm resulting from accidents during 2000 simulated scenarios.** The observed accidents essentially confirm the result of the risk distribution. (Fig. 5)

## Discussion

Ensuring the ethical behavior of AVs is a significant milestone in bringing them to public roads. Regulatory guidelines and recommendations have already been prepared for the question of unavoidable accidents. In line with these guidelines, we proposed an ethical algorithm for AV trajectory planning with a fair distribution of risk. Building our framework based on risks accounts for a suitable representation of the complex environment of AVs. Our developed ethical principles do not only focus on potentially hazardous situations but apply to all kinds of driving situations. They ensure transparent decision-making and, by considering responsibility, potentially account for long-term effects. Further, they may also be of interest for other AI applications beyond autonomous driving in the future.

While we provide a framework for ethical trajectory planning with our algorithm, we do not yet specify parameters such as the weighting factors of each principle or a value for the maximum acceptable risk. However, a particular parameter set may not apply universally. Different countries, cultures, or even individuals may demand different ethical emphases. Enabling user choice is an important requirement for including agile and continuous options.[13] The definition of these parameters should be the result of a social consensus agreed upon by all road users and susceptible to future research. To this end, the next step will be to conduct user studies with the parameter preferences of each user as a result. The findings of the user study can ultimately be used to validate the proposed risk distribution principles.

To validate our framework, we run simulations using an exemplary parameter set with equal weights for all principles. In contrast to selfish approaches, our fair risk distribution shows the intended behavior by shifting risks from VRUs to the autonomous ego vehicle. Granting VRUs a similar level of protection as other road users is one of the EU expert group's demands for AVs.[13] However, ethical decision-making requires valid risk modeling, which in this paper is simplified using only the uncertainties resulting from the prediction of other road users. For further development, it is important to account for possible bias in the modeling, which automatically influences ethical decision-making in our approach. However, the model cannot be arbitrarily accurate even without any modeling bias. Some extreme cases can only be inadequately covered. As Kauppinen[33] correctly notes, the determination of moral responsibility is arbitrarily complex. There is a difference between a pedestrian running into the street out of carelessness and someone being pursued and running for his life. For ethical decision-making, however, we must acknowledge the limited information that can reasonably be provided to the algorithm from a technical point of view[39]. Perfect all-encompassing solutions will probably never be achieved, and errors can also occur in ethical decision-making itself. But that should not stop us from starting to incorporate ethical considerations into the software of autonomous vehicles to idealize what best practices and its implications could be.

Finally, comparing the presented decision-making with that of a human driver reveals that human drivers do not perform this kind of risk-based reasoning but primarily act instinctively[31]. However, this should not serve as an argument against the integration of ethics into the software of AVs either: wouldn't it be nice to know that an AV not only drives more safely but also makes ethically-founded decisions?

## Data availability

All data gathered in this research are available in the Supplementary data file. This includes the evaliaton files for the simulated scenarios with various algorithms.

## Code availability

The algorithm for trajectory planning, as well as corresponding tools for analysis and visualization, are available open-source at:
*https://github.com/TUMFTM/EthicalTrajectoryPlanning*

## Fundig

The authors received financial support from the Technical University of Munich—Institute for Ethics in Artificial Intelligence (IEAI). Any opinions, findings and conclusions or recommendations expressed in this material are those of the authors and do not necessarily reflect the views of the IEAI or its partners.9